# GS_DravidianLangTech@2025: Women Targeted Abusive Texts Detection on Social Media

**Girma Yohannis Bade**[1,a], **Zahra Ahani**[2,a], **Olga Kolesnikova**,
**José Luis Oropeza**[4,a], **Grigori Sidorov**[5,a]

[a]Centro de Investigacion en Computación(CIC),
Instituto Politécnico Nacional(IPN), Miguel Othon de Mendizabal,
Ciudad de México, 07320, México.
[1]girme2005@gmail.com

## Abstract

The increasing misuse of social media has become a concern; however, technological solutions are being developed to moderate its content effectively. This paper focuses on detecting abusive texts targeting women on social media platforms. Abusive speech refers to communication intended to harm or incite hatred against vulnerable individuals or groups. Specifically, this study aims to identify abusive language directed toward women. To achieve this, we utilized logistic regression and BERT as base models to train datasets sourced from DravidianLangTech@2025 for Tamil and Malayalam languages. The models were evaluated on test datasets, resulting in a 0.729 macro F1 score for BERT and 0.6279 for logistic regression in Tamil and Malayalam, respectively.

## 1 Introduction

Abusive speech encompasses communication designed to inflict harm or foster hatred toward vulnerable individuals or groups, targeting attributes such as gender, ethnicity, religion, complexion, or physical abilities (Shahiki Tash et al., 2025; Kolesnikova et al., 2025). The use of offensive language in such speech can profoundly impact the psychological well-being of victims, potentially leading to adverse outcomes (Prasanth et al., 2022) such as causing psychological effects (Ahani et al., 2024a; Tash et al., 2024c), including lowering self-esteem and increasing anxiety(Li, 2022; Ramos et al., 2024b). On the other hand, with the rapid development of technology, traditional forms of communication such as newspapers, radio and television have been overtaken by Internet-based media platforms (Bade and Seid, 2018). These digital channels provide personal communication opportunities and are very accessible to people who previously did not have a platform to interact with a wider audience. As a result, social media has become an integral part of global communication, connecting people around the world (Sandaruwan et al., 2020), and this may enable distributing misogynistic content to the corner of the world within a minute.

However, there are efficient ways to automate the detection of this content using machine learning and deep learning models (Arif et al., 2024; Ramos et al., 2024a), with advanced approaches such as transformer architectures and large language models (LLM) (Tash et al., 2024b) showing significant efficiency in identifying and analyzing linguistic patterns (Vaswani, 2017; Ahani et al., 2024b). Social media platforms have become vital arenas for discussing violence against women (VAW), with movements like #MeToo raising global awareness about sexual harassment and abuse. However, these same platforms often expose women to targeted harassment, including abusive language and threats, particularly against activists, public figures, and those in leadership roles. This online abuse not only perpetuates harmful societal norms but also limits women's participation in public and digital spaces. Efforts such as bystander intervention, digital activism, and community-level education have been instrumental in challenging these norms, though resource constraints and recurring harassment underscore the need for stronger institutional support to foster safer digital environments (efore #MeToo: Violence against Women on Social Media (Jenkins and Nickerson, 2019).

TThe objective of our research is to detect abusive remarks in Tamil and Malayalam, two Dravidian languages that are primarily used in South Asia and are acknowledged as one of India's 22 scheduled languages (Priyadharshini et al., 2022a). However, the limited availability of linguistic resources for Tamil poses significant challenges for natural language processing (NLP), especially in building robust datasets. The detection of offensive comments involves text classification, where the goal is to classify the text into predefined classes (Priyadharshini et al., 2023; Zamir et al., 2024). While transformer models(Vaswani, 2017; Tash et al., 2024a) have been effectively employed for this purpose, the language limitations of Tamil have limited the exploration of techniques such as data augmentation and have not developed this field.

We aimed to determine the most efficient model by thoroughly analyzing the results, emphasizing the critical need to address the challenge of reducing offensive comments against women in Tamil and Malayalam language online communities.

## 2 Related work

Several studies have explored detecting online misogyny using Natural Language Processing (NLP) (Bade et al., 2024a). Research has focused on challenges such as informal language, code-mixing (Bade et al., 2024a), and context-dependent expressions of misogyny (Bade et al., 2024a). Notably, a study by (Li, 2022)introduced a misogyny detection system using deep learning models trained on large datasets of online comments. Another key contribution is (Strathern and Pfeffer, 2022), which utilized transformer-based models to detect gender-based hate speech across multiple languages. Despite progress, detecting misogyny in low-resource languages remains underexplored (Bade et al., 2024c), with limited datasets and resources for languages like Tamil and Malayalam, which highlights the need for culturally sensitive and linguistically diverse approaches to tackle online misogyny.

Detecting abusive content online is challenging, largely because of the subtle distinction between offensive language and free speech (Khairy et al., 2021).

Efficient and precise detection and classification of abusive comments are essential for creating safer and more inclusive online platforms and communities(Bansal et al., 2022; Priyadharshini et al., 2022b). The study (Chakravarthi et al., 2023) introduced four Tamil and Tamil-English datasets from YouTube, addressing challenges like informal language and code-mixing. While MURIL excelled in binary abuse detection, fine-grained tasks favored classical models due to limited data. This was the first Tamil-focused FGACD study, providing datasets and methods for low-resource languages. (Roy et al., 2022)emphasize the urgent necessity of promptly identifying and eliminating hate speech and offensive content from social media to prevent its harmful impact and rapid dissemination. They point out that code-mixed languages like Hindi–English, Tamil–English, Malayalam–English, and Telugu–English make it especially difficult to identify hate speech. Their study provides a detailed comparison of various machine learning and deep learning approaches to tackle this problem.

The study (Shanmugavadivel et al.)demonstrates that in low-resource languages like Tamil, adapter-based approaches outperform fine-tuned models. Additionally, the hyperparameter optimization framework Optuna is utilized to identify optimal hyperparameter values for improved classification performance. Among the proposed models, MuRIL (Large) achieves the highest performance with an accuracy of 74.7%, surpassing other models tested on the same dataset.

## 3 Task description

This task focuses on the detection of abusive text targeting women on social media, specifically in Tamil and Malayalam, two low-resource Dravidian languages. Social media platforms, while fostering communication and entertainment, have also become spaces where women face gender-based abuse, including hateful, derogatory, or threatening comments. This form of online harassment reflects societal biases and poses significant psychological, social, and professional challenges to the victims.

This task aims to identify and classify comments as either Abusive or Non-Abusive. The dataset used for this task was curated by scraping YouTube comments, focusing on controversial and sensitive topics where gender-based abuse is prevalent. These comments include a range of abusive language types, such as explicit abuse, implicit biases, stereotypes, and coded language.

The annotated datasets are provided to participants in both Tamil and Malayalam. Participants are required to build and evaluate models that can accurately classify each comment into one of the predefined categories: Abusive Non-Abusive.

## 4 Methodology

### 4.1 Datasets

The dataset comprises text samples categorized into two classes: Non-Abusive and Abusive. It includes 1,424 Non-Abusive entries and 1,366 Abusive entries in the Tamil training dataset, providing a nearly balanced distribution. Similarly for Malayalam, 1424 are Non-abusive and 1366 Abusive are there in training. This data set is well suited for training and evaluating models for binary classification tasks, such as detecting abusive language in text. Its balance ensures fair representation of both classes, enabling robust model performance analysis.

| Languages | Dataset | has_label? | Size |
|---|---|---|---|
| Tamil | Train | yes | 2790 |
| | Dev | yes | 598 |
| | Test | no | 597 |
| Malayalam | Train | yes | 2933 |
| | Dev | yes | 629 |
| | Test | no | 628 |

Table 1: Dataset statistics

### 4.2 Dataset Preprocessing:

Preprocessing steps included: removing URLs, special characters,redundant spaces, and maintaining inconsistencies. Mapping class labels to binary values: "Abusive" (1) and "Non-Abusive" (0).

### 4.3 Expriment Models

**BERT (Bidirectional Encoder Representations from Transformers)**

BERT was employed for Tamil language in this study. It captures contextual information bidirectionally, making it effective for fine-tuning for binary classification with two output labels.

Tokenization and feature extraction for BERT transformer models was performed using their respective BertTokenizer (Bade et al., 2024b). The hyperparameters used for BERT model are summarized in the following Table 2. It's crucial to modify the hyperparameters in machine learning. The objective is to determine which parameter values result in the best model accuracy. The hyperparameters that each classifier possesses can have a big impact on how well the classifier performs (Gomes et al., 2023).

| Hyperparameters | Values |
|---|---|
| Learning Rate | 1e-5 |
| Evaluation Strategy | epoch |
| Epochs | 5 |
| Batch Size | 32 |
| Max length | 128 |
| Activation function | sigmoid |

Table 2: BERT Hyperparameters

As we can see from Table 2, The learning rate shows how many executions are made to enhance the model's functionality. An epoch is when the learning algorithm goes over the full training dataset once (Mersha et al., 2024). Thus, we set the epoch to be 5,.i.e the execution did pass 5 complete times. The batch size refers to dividing the total data size into 32 and bringing the divided batch one a time for the execution. This helps the execution to be fast. The last parameter, activation function is used to label or group the computational result into two classes.

**Logistic Regression (LR)**: we implemented this approach to Malayamal langauge's data. It's a classic machine learning algorithm that predicts the relationship between input variables and the likelihood that an output will belong to a particular class (Krasitskii et al., 2024, 2025). Its sigmoid function transforms linear combinations of input features into probabilities, mapping real-valued predictions to a range between 0 and 1. This probabilistic output is critical for classification tasks, as it enables a threshold-based decision-making process. Since it has no its own feature extracting tool, we employed the TF-IDF (Term Frequency Inverse Document Frequency) feature extracting technique to tokenize and convert text data into numeric form (Bade et al., 2024c). For text feature extraction, the TF-IDF technique uses word statistics. This ignores the possibility that words could be represented by their synonyms and only takes into account ways that they are used consistently throughout all texts, like ASCLL (Liu et al., 2018). Table 3 summarizes the selected models and their feature extraction techniques.

| Language | Modles | Feature Extraction |
|---|---|---|
| Tamil | BERT | BertTokenizer |
| Malayamal | LR | TF-IDF |

Table 3: Model selection and their respecive feature extraction techniques

Figure 1 illustrates the proposed system's architecture. The process begins with dataset preprocessing to ensure data quality, followed by feature extraction, converting text into numerical form for model training. BERT undergoes fine-tuning, while LR trains from scratch to distinguish abusive and non-abusive content in Tamil and Malayalam. The trained model is then tested, generating predictions, which are evaluated against manually labeled test data using macro F1 to assess performance.

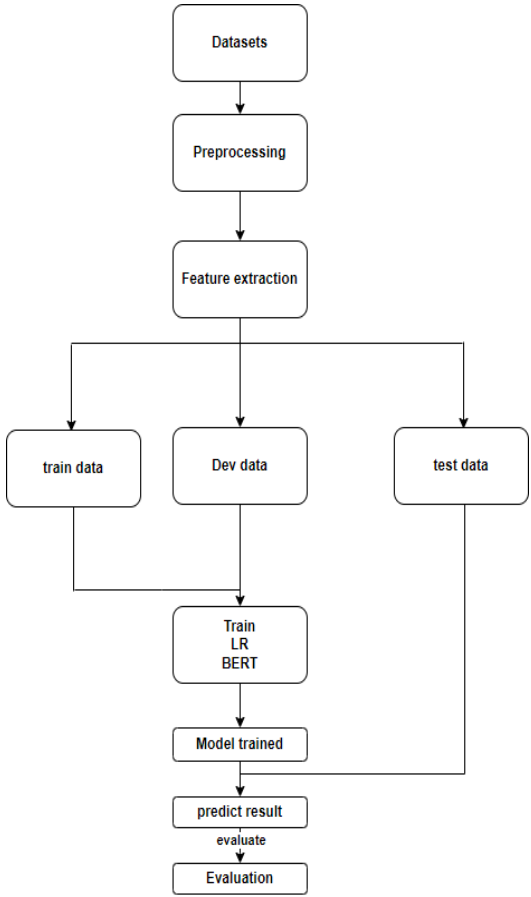

Figure 1: Proposed system archtecture

## 5 Result and Discussion

The results represent a significant and exciting milestone in any experimental researches. As we trained the provided dataset on the chosen algorithm and tested its performance using a separate test set, the predictions generated from the trained models were submitted to the workshop organizers for evaluation. These submissions were assessed using the evaluation metric of macro F1 score. The final results, when published by the organizers, reveals that BERT attained a significantly higher macro F1 score of 0.7293 for Tamil, whereas for Malayalam, LR attained a macro F1 score of 0.6279. Table 4 shows more details.

Table 4: Macro F1-scores of BERT model and LR on two languages' data

| **Language** | **Model** | **Macro F1-Score** |
|---|---|---|
| Tamil | BERT | 0.7293 |
| Malayalam | LR | 0.6279 |

As we understand from the result Table 4, the model bert has resulted impressive prediction for this particular task,especially for Tamil. Figure 2 and 3 illustrate the results in visual manner using confusion matrix.

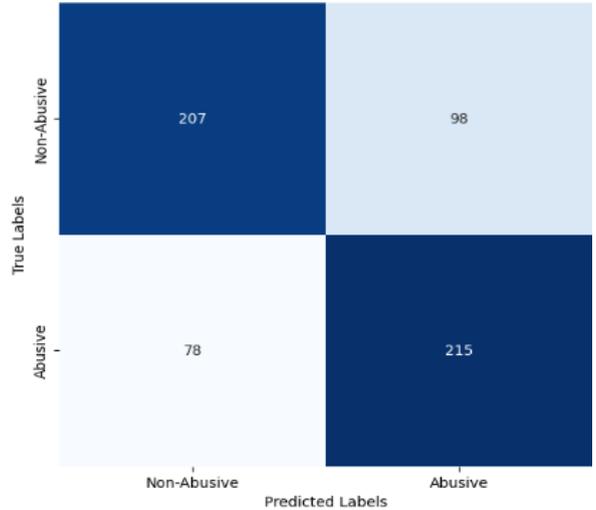

Figure 2: The confusion matrix that shows the bert model predictions on Tamil.

In Figure 2, our developed model predicted 207 instances as non-abusive, and they are actually too. Similarly, the model predicted 215 instances Abusive, and they are actually abusive too. On the other hand, there were two errors. By mistake, our model predicted 78 as abusive, but they are actually non-abusive, and it also predicted 98 instances as non-abusive, but actually it's abusive.

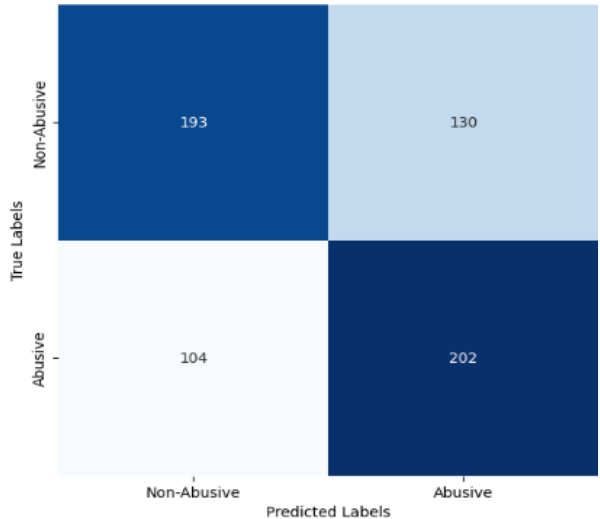

Figure 3: The confusion matrix that shows the logistic regression's model predictions on Malayamal.

In the Figure 3, our developed model predicted 193 instances as non-abusive, and they are actually too. Similarly, the model predicted 202 instances of abuse, and they are actually abusive too. In the other hand, two errors occurred. Our model, by mistake, predicted 104 as abusive, but they are actually non-abusive, and it also predicted 130 instances as non-abusive, but actually they are abusive instances.

## 6 Conclusion and Future Work

In this task, we have developed two models to classify Tamil and Malayamal social media comments. We built the model for Tamil data based on BERT and for Malayamal based on Logistic Regression. The models are evaluated their conformance using various metrics. They are able to classify test dataset into into 'Abusive' and 'Non-Abusive' as expected. As the results show, bert outperformed Logistic Regression in this usecase.

Since social media posts are useful to identify the political opinion, the jobs ought to be transferred into other various languages. Furthermore, by offering additional algorithms for the languages utilized here and expanding the number of dataset sizes, the performance of the suggested model in this study should be enhanced.

## Limitation and Ethics Statement

The dataset of language Tamil and Malayamal here were provided with limited data size and the model was trained using a this tiny dataset. Hence, the performance observed may not generalize well to all unseen data. However, our model has demonstrated a comparable performance in detecting women targeted abusive comments for social media posts. Furthermore, our work adheres to the ethical principles outlined for computational research and professional conduct[1].

## Acknowledgements

The work was done with partial support from the Mexican Government through the grant A1-S-47854 of CONACYT, Mexico, grants 20241816, 20241819, and 20240951 of the Secretaría de Investigación y Posgrado of the Instituto Politécnico Nacional, Mexico. The authors thank the CONACYT for the computing resources brought to them through the Plataforma de Aprendizaje Profundo para Tecnologías del Lenguaje of the Laboratorio de Supercómputo of the INAOE, Mexico and acknowledge the support of Microsoft through the Microsoft Latin America PhD Award.

---

[1]https://www.aclweb.org/portal/content/acl-code-ethics